\documentclass[10pt,twocolumn,letterpaper]{article}

\usepackage{cvpr}              %

\usepackage{graphicx}
\usepackage{amsmath}
\usepackage{amssymb}
\usepackage{booktabs}
\usepackage{bbding} 

\usepackage[table]{xcolor} 
\definecolor{best1}{RGB}{222,242,212}
\definecolor{best2}{RGB}{255,250,212}

\usepackage[pagebackref,breaklinks,colorlinks]{hyperref}

\usepackage[capitalize]{cleveref}
\crefname{section}{Sec.}{Secs.}
\Crefname{section}{Section}{Sections}
\Crefname{table}{Table}{Tables}
\crefname{table}{Tab.}{Tabs.}

\def\confName{CVPR}
\def\confYear{2022}

\newcommand{\minye}[1]{}  %

\begin{document}

\title{Fourier PlenOctrees for Dynamic Radiance Field Rendering in Real-time}

\author{Liao Wang$^{1*}$
\and
Jiakai Zhang$^{1*}$
\and
Xinhang Liu$^1$
\and
Fuqiang Zhao$^1$
\and
Yanshun Zhang$^{3}$
\and
Yingliang Zhang$^{3}$
\qquad \qquad
Minye Wu$^{2}$ \qquad \qquad Lan Xu$^1$ \qquad \qquad Jingyi Yu$^1 $\\
$^{1}$ ShanghaiTech University \qquad $^{2}$ KU leuven \qquad $^{3}$ DGene \\
}
\maketitle

\begin{abstract}
     Implicit neural representations such as Neural Radiance Field (NeRF) have focused mainly on modeling static objects captured under multi-view settings where real-time rendering can be achieved with smart data structures, e.g., PlenOctree. 
	In this paper, we present a novel Fourier PlenOctree (FPO) technique to tackle efficient neural modeling and real-time rendering of dynamic scenes captured under the free-view video (FVV) setting.
	The key idea in our FPO is a novel combination of generalized NeRF, PlenOctree representation, volumetric fusion and Fourier transform.
	To accelerate FPO construction, we present a novel coarse-to-fine fusion scheme that leverages the generalizable NeRF technique to generate the tree via spatial blending. To tackle dynamic scenes, we tailor the implicit network to model the Fourier coefficients of time-varying density and color attributes.
	Finally, we construct the FPO and train the Fourier coefficients directly on the leaves of a union PlenOctree structure of the dynamic sequence. 
	We show that the resulting FPO enables compact memory overload to handle dynamic objects and supports efficient fine-tuning.
	Extensive experiments show that the proposed method is 3000 times faster than the original NeRF and achieves over an order of magnitude acceleration over SOTA while preserving high visual quality for the free-viewpoint rendering of unseen dynamic scenes.

\end{abstract}

{\let\thefootnote\relax\footnote{* Authors contributed equally to this work. }}\par
\section{Introduction}
\label{sec:intro}

Interactive and immersive applications, such as Telepresence and Virtual Reality (VR), make plenty use of free-viewpoint videos to provide unique and fully controllable viewing experiences. At the core are fast generation and real-time rendering at new viewpoints with ultra-high photorealism. Traditional image-based modeling and rendering approaches rely on feature matching and view interpolation, whereas the latest neural rendering techniques are able to integrate the two processes into a deep net that simultaneously represents the geometry and appearance for efficient rendering. By far, most neural rendering techniques have focused on modeling static objects and employing smart data structures. For example, volumetric neural modeling techniques ~\cite{lombardi2019neuralvol, mildenhall2020nerf} overcome many limitations of traditional methods, including tailored matching algorithms and optimization procedures and can even tackle non-Lambertian materials. 
The seminal work of the Neural Radiance Field (NeRF) ~\cite{mildenhall2020nerf} learns a neural representation based on MLP to represent static scenes as radiance fields with the property of density and color. It only requires calibrated multi-view images to produce compelling free-viewpoint rendering. However, the MLP structure is still too slow to achieve real-time performance. Existing techniques explore using thousands of tiny MLPs~\cite{reiser2021kilonerf}, applying factorization~\cite{garbin2021fastnerf}, tailored volumetric data structures~\cite{yu2021plenoctrees, hedman2021baking}, and primitive-based rendering~\cite{10.1145/3450626.3459863}. Despite their effectiveness, very few techniques are directly applicable to handle dynamic scenes, in particular, objects with non-rigid deformations such as the human body. In this work, we present a novel neural representation for generating free-viewpoint videos from multi-view sequence inputs as well as for real-time photorealistic rendering.

\begin{figure}[t]
	\begin{center}
		\includegraphics[width=1.0\linewidth]{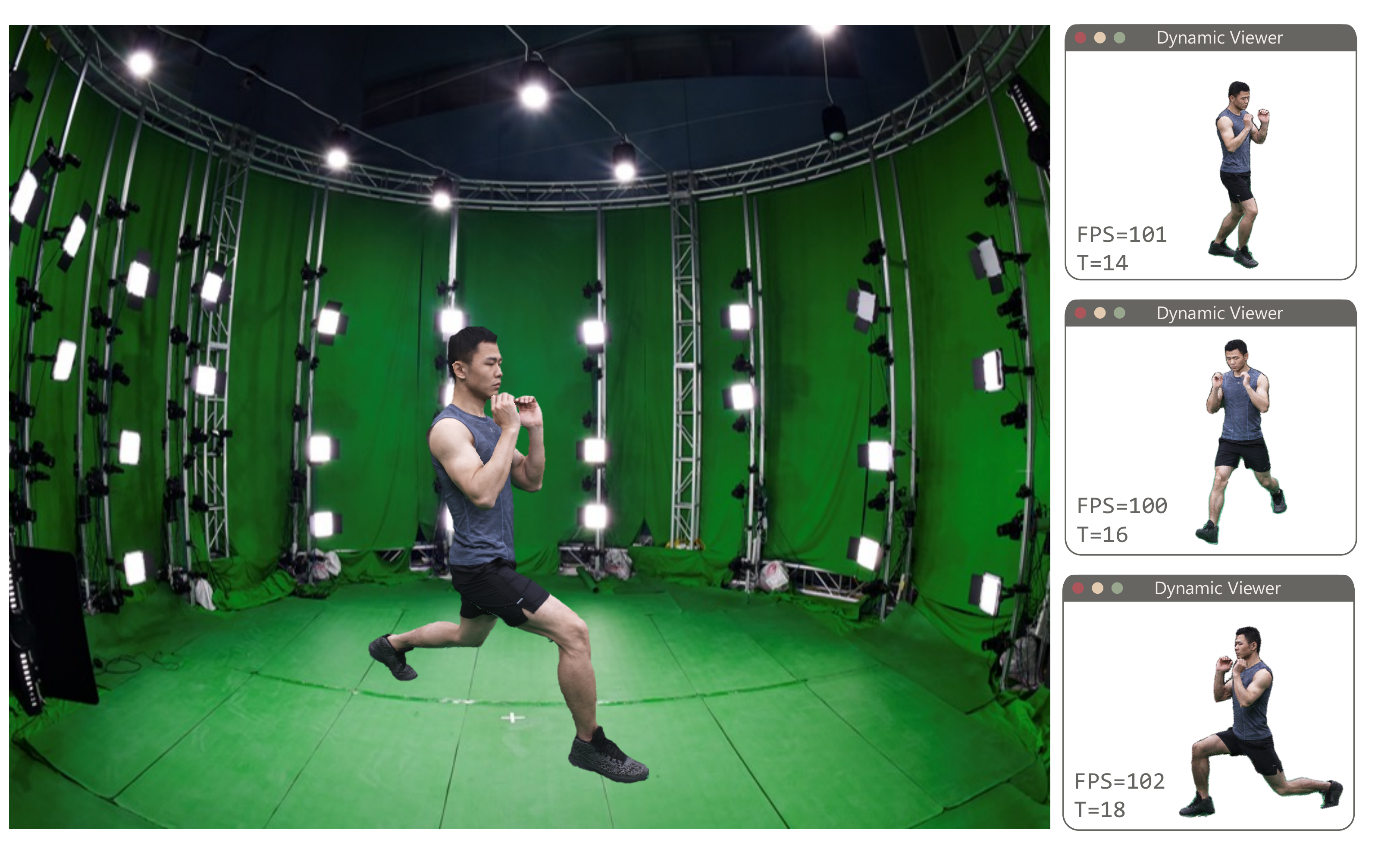}
	\end{center}
	\vspace{-0.5cm}
	\caption{Our method achieves a photo-realistic rendered result for dynamic scenes in real-time based on our novel Fourier PlenOctree structure.}
	\label{fig:teaser}
	\vspace{-7mm}
\end{figure}

Several recent efforts have investigated combining neural modeling with classical flow or geometry estimations ~\cite{pumarola2021d, park2020deformable}. For example, one can apply motion flows to explicitly transform sampled points in individual frames to a canonical model to partially account for non-rigid deformations. However, they are vulnerable to lost tracks and can lead to motion artifacts. There are also emerging interests on utilizing pre-defined (e.g. skeleton~\cite{peng2021animatable} or parametric models~\cite{liu2021neural, peng2021neural}) to explicitly calculate stable motion flows from model animations. These approaches, however, are limited to handling specific types of objects consistent with the pre-defined model. \cite{wang2021ibutter} directly predicts a neural radiance field using a general network for each frame while avoiding online training. Its rendering speed, however, is not yet sufficient for interactive and immersive experiences.

In this paper, we present a novel Fourier PlenOctree (FPO) technique for neural dynamic scene representation, which enables efficient neural modeling and real-time rendering of unseen dynamic objects with compact memory overload, as illustrated in Fig.~\ref{fig:teaser}. The key idea in our FPO is a novel combination of generalized NeRF, PlenOctree representation, volumetric fusion, and Fourier transform.

For efficient scene modeling, we present a novel coarse-to-fine fusion scheme that leverages generalizable NeRF~\cite{wang2021ibrnet} technique to generate the PlenOctree with fast plenoptic functions inference.
Inspired by the volumetric fusion strategy~\cite{KinectFusion}, we propose a spatial blending scheme to generate the neural tree in the order of minutes.
To tackle dynamic scenes, we tailor the implicit network to model the Fourier coefficients of time-varying density and plenoptic functions of the dynamic PlenOctree.
By discarding high-frequency bases, our novel representation can achieve high storage efficiency while persevering perceptive details.   
Finally, we construct the FPO and train the Fourier coefficients directly on the leaves of a union PlenOctree structure of the dynamic sequence. 
By combining the benefits of PlenOctree rendering and Fourier operations, our FPO enables real-time free-viewpoint synthesis of dynamic scenes and supports efficient fine-tuning. Comprehensive experiments show that FPO is 3000 times faster than the original NeRF implementation and achieves over an order of magnitude acceleration over state-of-the-art techniques for dynamic scene modeling and rendering.
To summarize, our contributions include:
\begin{itemize}
    \item We introduce a FPO representation that enables real-time rendering of general dynamic scenes with fast fine-tuning and compact memory overload.
    
    \item We present a coarse-to-fine scheme  that utilizes
    generalizable NeRF for PlenOctree generation and constructing FPO efficiently.

\end{itemize}

\begin{figure*}[th]
\begin{center}
    \includegraphics[width=\linewidth]{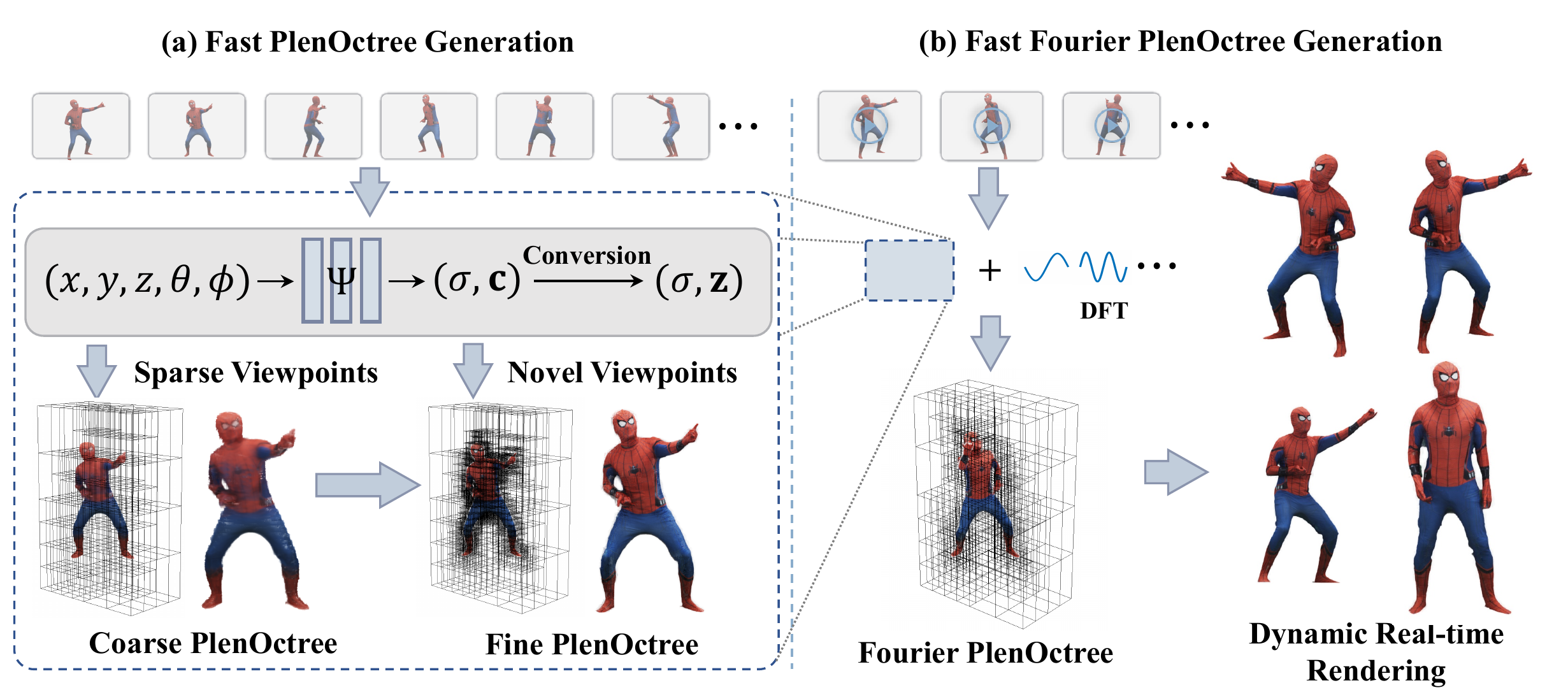}
\end{center}
\vspace{-5mm}

\caption{Illustration of our fast PlenOctree generation for static scene and fast Fourier PlenOctree generation for dynamic scene. (a) illustrates how to generate a PlenOctree from multiview images. Given these images, a generalized NeRF $\Psi$ predicts view-dependent density $\sigma$ and color $\mathbf{c}$ by inputting 3D sample point $(x, y, z)$ with view direction $(\theta, \phi)$, then we can convert them to view-independent density $\sigma$ and SH coefficients $\mathbf{z}$. Using sparse view RGB images and masks rendered by $\Psi$, we can obtain a coarse PlenOctree. Finally we fine-tune it to be a fine Plenoctree by inputting dense view images rendered by $\Psi$. (b) extends the pipeline to dynamic scene by combining the idea with Discrete Fourier Transform(DFT) and achieves a fast Fourier PlenOctree generation and real-time rendering for dynamic scene using Fourier PlenOctree.}
\label{fig:pipeline}
\vspace{-7mm}
\end{figure*}

\section{Related work}
\label{sec:Relatedwork}

\textbf{Novel View Synthesis for Static Scenes. }
The task of synthesizing novel views of a scene given a set of photographs has attracted the attention of the community.  All current methods predict an underlying geometric or image-based 3D representation that allows rendering from novel viewpoints. 

Among them, mesh-based representations \cite{waechter2014let, wood2000surface} are compact and easy to render; however, optimizing a mesh to fit a complex scene of arbitrary topology is challenging.  Volume rendering is a classical technique with a long history of research in the graphics community \cite{drebin1988volume}. Volume-based representations such as voxel grids \cite{seitz1999photorealistic,kutulakos2000theory} and multi-plane images (MPIs) \cite{penner2017soft, szeliski1998stereo} are a popular alternative to mesh representations due to their topology-free nature: gradient-based optimization is therefore straightforward, while rendering can still be real-time. 

The most notable approach Neural Radiance Field (NeRF)~\cite{mildenhall2020nerf} combines the implicit representation with volumetric rendering 
to achieve compelling novel view synthesis with rich view-dependent effects.
However, these neural representations above can only handle static scenes. %

\textbf{Novel View Synthesis for Dynamic Scenes. }
Different from static scenes, tackling dynamic scenes requires settling the illumination changes and moving objects. 
One approach is to obtain a reconstruction for dynamic objects with input data from either camera array or a single camera. Methods involving silhouette~\cite{taneja2010modeling, kim2010dynamic}, stereo~\cite{luo2020consistent, li2019learning, lv2018learning,FlyCapLan}, segmentation~\cite{russell2014video, ranftl2016dense}, and photometric~\cite{ahmed2008robust,vlasic2009dynamic,he2021challencap} have been explored. 
Early solutions~\cite{Mustafa_2016_CVPR,collet2015high,motion2fusion} rely on multi-view dome-based setup for high-fidelity reconstruction and texture rendering of human activities in novel views.
Recently, volumetric approaches with RGB-D sensors and modern GPUs have enabled real-time novel view synthesis for dynamic scenes and eliminated the reliance on a pre-scanned template model.
The high-end solutions~\cite{dou-siggraph2016,motion2fusion,TotalCapture,UnstructureLan} rely on multi-view studio setup to achieve high-fidelity reconstruction and rendering, while the low-end approaches~\cite{newcombe2015dynamicfusion,FlyFusion,robustfusion} adopt the most handy monocular setup with a temporal fusion pipeline~\cite{KinectFusion} but suffer from inherent self-occlusion constraint.

Recent work~\cite{park2020deformable,pumarola2021d,li2020neural,xian2020space,tretschk2020non,rebain2020derf,ost2020neural,st-nerf} extend the approach NeRF~\cite{mildenhall2020nerf} using neural radiance field into the dynamic settings. 
They decompose the task into learning a spatial mapping from the current scene to the canonical scene at each timestamp and regressing the canonical radiance field. 
However, the above solutions using dynamic neural radiance fields still suffer from a long training time as well as rendering time.

\textbf{NeRF Accelerations. }While NeRFs can produce high-quality results, their computationally expensive rendering leads to slow training and inference. One way to speed up the process of fitting a NeRF to a new scene is to incorporate priors learned from a dataset of similar scenes. This can be accomplished by conditioning on predicted images features~\cite{trevithick2020grf,yu2020pixelnerf,wang2021ibrnet} or meta-learning~\cite{tancik2020meta}. To improve rendering speed, Neural Sparse Voxel Fields (NSVF)~\cite{nsvf} learns sparse voxel grids of features that are input into a NeRF-like model. The sparse voxel grid allows the renderer to skip over empty regions when tracing a ray which improves the rendering time $\sim$10x. AutoInt~\cite{lindell2020autoint} modifies the architecture of the NeRF so that inference requires fewer samples but produces lower quality results. 

NeX~\cite{wizadwongsa2021nex} extends MPIs to encode spherical basis functions that enable view-dependent rendering effects in real-time. \cite{hedman2021snerg,garbin2021fastnerf,reiser2021kilonerf} also distill NeRFs to enable real-time rendering. \cite{yu2021plenoctrees} use an octree-based 3D representation which supports view-dependent effects to achieve real-time performance.

However, none of the current methods tackles the challenge to accelerate the training and rendering process of the dynamic radiance field.

\section{Generalized PlenOctree Fusion}\label{sec:3}

Recall that NeRF takes an MLP as a mapping function to predict a density $\sigma$ and a color $\mathbf{c}$ for a queried 3D point $p = (x,y,z)$ in a given viewing direction $\mathbf{d} = (\theta,\phi)$. 
To boost the NeRF rendering procedure, \cite{yu2021plenoctrees} modifies the outputs of the mapping function to Spherical Harmonic (SH) coefficients $ \mathbf{z} \in \mathbb{R}^{\ell_{max}^2\times3}$ with a density $\sigma$, which will be cached in leaves of PlenOctree as an initialization. 
Having $\mathbf{z}$ and $\sigma$, we can calculate the color of queried 3D point in given viewing direction:
\begin{equation}
    \mathbf{c}(\mathbf{d};\mathbf{z}) = S\left( \sum^{\ell_{max}}_{\ell=0}\sum^{\ell}_{m=-\ell}z_{m,\ell}Y_{m,\ell}(\mathbf{d}) \right)
\end{equation}
where $S$ is the sigmoid function to normalize color, $Y_{m,\ell}: S^2 \rightarrow \mathbb{R}$ is a real-valued basis function of SH. 

Even though the rendering speed of PlenOctree is rather fast due to this simple calculation, the acquisition of SH coefficients and densities is still time-consuming.  
Therefore, we present a novel  coarse-to-fine fusion scheme that leverages the generalizable NeRF technique $\Psi$~\cite{wang2021ibrnet,wang2021ibutter,chen2021mvsnerf} to attack this problem via spatial blending.
In the following, we demonstrate PlenOctree Fusion algorithm in traditional static scenes as an example. 
Note that we can also do PlenOctree Fusion in Fourier PlenOctree to deal with dynamic scenarios in the same way since both of them have very similar data structures which will be introduced in Sec.~\ref{sec:4.2}.

A generalized neural rendering network $\Psi$ takes images of adjacent views of a target view as inputs and infers an implicit volume with regard to the target view.
We can directly query colors and densities of sample points corresponding to leaves in PlenOctree from this volume without per-scene training. 
However, these colors and densities are all with respect to a specific target view due to different view directions. 
To obtain a completed PlenOctree, we need to sample more target views and fuse their local PlenOctree together. 
Fig.~\ref{fig:pipeline} (a) illustrates our pipeline.

The proposed PlenOctree Fusion follows a coarse-to-fine strategy. 
Specifically, to obtain coarse PlenOctrees, we initialize a coarse PlenOctree with $N^3$ voxel grids as tree leaves. 
Given multi-view images and silhouettes extracted via chroma key segmentation and background subtraction, $\Psi$ predicts images for 6 sparse views which are uniformly around the object. 
Then Shape-From-Silhouette (SFS) method~\cite{baumgart1974geometric} is applied to generate a coarse visual hull from sparse view silhouettes. 
For each leaf inside the visual hull, we uniformly sample directions $\theta,\phi \in [0, 2\pi]$ to predict densities and colors by feeding positions and directions to $\Psi$. 
Note that predicted densities and colors are both view-dependent, denoted as $\sigma(\theta,\phi)$ and $\mathbf{c}(\theta,\phi)$ respectively. 
Next, we need to convert those predictions to view-independent densities and SH coefficients $\mathbf{z}$ for each leaf by:
  \begin{equation}
\begin{split}
    \sigma &= \mathbb{E}({\sigma}(\theta, \phi))\\
    \mathbf{z} &= \mathbb{E}(\text{SH}(\theta,\phi)\cdot\mathbf{c}(\theta,\phi)),
\end{split}
\end{equation}
where $\mathbb{E}(\cdot)$ is the expectation operator, and SH$(\theta, \phi)$ is an evaluation function which calculates SH from a direction. 
After filling in all leaves, we obtain a coarse PlenOctree.

As the coarse PlenOctree tree is generated from sparse viewpoints, many redundant leaves need to be filtered out. Also, the values of leaves are not accurate. 
In the fine stage, inspired by fusion-based methods\cite{KinectFusion, gao2019dynamic}, we first render 100 dense view images by PlenOctree, and query the points that the transmittance in volume rendering $T_i > 1e-3$ from $\Psi$, then we fuse PlenOctree using these points by calculating the updating weights for each leaf.  The reason why PlenOctree is initialized from 6 views is that the 6 views query all the points, while the 100 views will only query about 1\% points which are fast compared to querying all the points of 100 views.
At $i$-th viewpoints, we use the transmittance $T_i(x,y,z)$ as update weight for leaves at position $(x,y,z)$ and update density and SH coefficients by the following equation:
\begin{equation}
    \sigma_i =
    \dfrac{W_{i-1}\sigma_{i-1}+T_i\sigma_i}{W_{i-1}+T_i}
\end{equation}

\begin{equation}
    \mathbf{z} =
    \dfrac{W_{i-1}\mathbf{z}_{i-1}+T_i\cdot\text{SH}(\theta_i,\phi_i)\cdot\mathbf{c}(\theta_i,\phi_i)}{W_{i-1}+T_i}
\end{equation}
Then calculate weight and count of updated times:
\begin{equation}
    \begin{aligned}
    W_{i} = \frac{C_{i}-1}{C_{i}} W_{i-1} + \frac{1}{C_{i}}T_i
    \end{aligned}
\end{equation}
where $C_i = C_{i-1} + 1$ means how many times the leaf has been updated at $i$-th step. 
After these iterations, we filter out the leaves which have $\sigma < 1e-3$ to save the PlenOctree storage and further computational costs. 
Using PlenOctree fusion with $\Psi$ instead of naive gradient descend to do updating can avoid a tremendous amount of optimization iterations so as to accelerate refinement.

With the help of the generalized PlenOctree Fusion, we can obtain a PlenOctree representation for a static scene within 60 seconds, which greatly improves generation speed.

\begin{figure}[t]
\begin{center}
    \includegraphics[width=\linewidth]{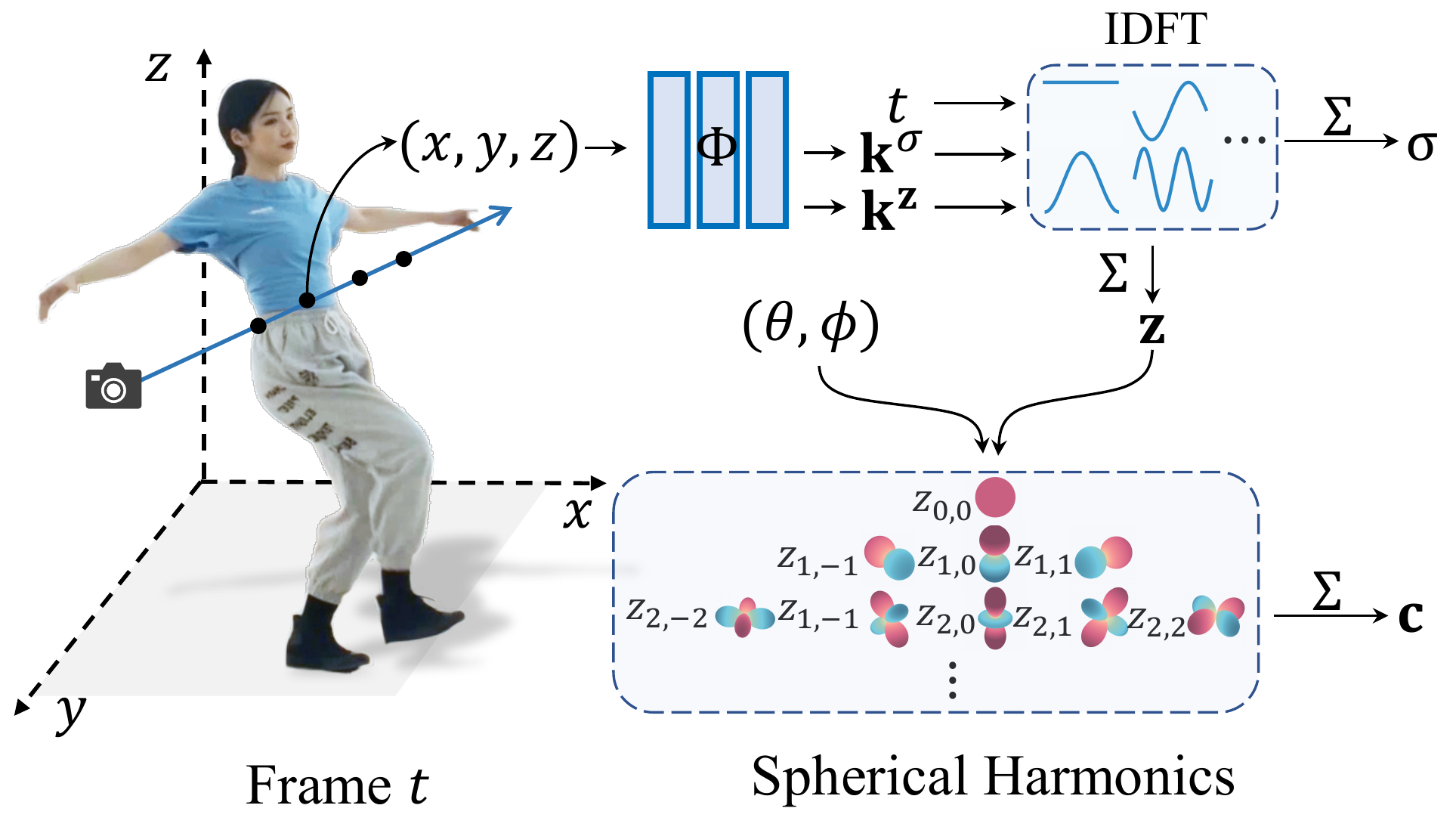}
\end{center}
\vspace{-6mm}
\caption{Illustration of our 4D Scene Representation in frequency domain, including first predicting Fourier coefficients $\mathbf{k^\sigma}$ and $\mathbf{k^z}$ by inputting $(x,y,z)$, then computing density $\sigma$ and factors $\mathbf{z}$ of SH basis by summing the weighted Fourier Transform with additional timestamp $t$, finally, computing color $c$ by summing the weighted SH bases with viewing direction $(\theta,\phi)$.}
\label{fig:fig3}
\vspace{-6mm}
\end{figure}
\section{Fourier PlenOctree}\label{sec:4}

In this section, we elaborate how Fourier PlenOctree records and renders free-viewpoint videos. 
As illustrated in Fig.~\ref{fig:pipeline} (b), we introduce Fourier PlenOctree with a novel 4D scene representation, which adopts PlenOctree to dynamic scenes  by compressing time-variant information in the frequency domain (Sec.~\ref{sec:4.1}). 
Combined with Generalized PlenOctree Fusion, Fourier PlenOctree exhibits fast generation and real-time rendering abilities (Sec.~\ref{sec:4.2}). 
Fourier PlenOctree fine-tuning can further improve rendering quality within a short time (Sec.~\ref{sec:4.3}).

\subsection{4D Scene Representation in Frequency Domain}\label{sec:4.1}
We propose a novel 4D Scene Representation in a high dimensional frequency domain to enable efficient neural modeling and real-time rendering for general dynamic scenes.

As illustrated in Fig.~\ref{fig:fig3}, given a 4D scene sample point $(x, y, z, t)$, the mapping function $\Phi$ is defined as below:

\begin{equation}
\label{eqn:representation}
    \Phi(x, y, z) = \mathbf{k}^\sigma, \mathbf{k^z}
\end{equation}
where $\mathbf{k^\sigma} \in \mathbb{R}^{n_1}$ and $\mathbf{k^z} \in \mathbb{R}^{n_2 \times (\ell_{max}+1)^2 \times 3}$ are two Fourier Transform coefficients of the functions of density $\sigma(t)$ and SH coefficients $\mathbf{z}(t)$ at position $(x,y,z)$ respectively; 
$n_1$ is the Fourier coefficient number of $\sigma$, $n_2$ is the Fourier coefficient number of each SH coefficient $\mathbf{z}$,  note that $(\ell_{max}+1)^2 \times 3$ is the number of SH coefficients for RGB channels. 
As the timestamp $t$ is given, density $\sigma$ can be recovered by the following real-valued Inverse Discrete Fourier Transform (IDFT) in Eq.~\ref{eqn:idft_sigma}:

\begin{equation}
    \label{eqn:idft_sigma}
    \sigma(t;\mathbf{k^\sigma}) = \sum_{i=0}^{n_1-1} \mathbf{k^\sigma_i} \cdot \text{IDFT}_i(t)\\
\end{equation}
where $t$ is the frame index and 
\begin{equation}\label{eqn:IDFT}
    \text{IDFT}_i(t) = \left\{
    \begin{aligned}
    &\cos(\frac{i\pi}{T} t) &\text{ if $i$ is even}\\
    &\sin(\frac{(i+1)\pi}{T} t) &\text{ if $i$ is odd}\\
    \end{aligned}
    \right.
\end{equation}
To handle view-dependent effects, We use a similar idea to compute Fourier coefficients for each element $z_{m,l} \in \mathbb{R}^3 $ of coefficients $\mathbf{z} = (z_{m,l})^{m:-\ell\leq m \leq \ell}_{l: 0\leq \ell \leq \ell_{max}} $ of SH function by the following Eq.~\ref{eqn:idft_z}:

\begin{equation}\label{eqn:idft_z}
    z_{m,\ell}(t;\mathbf{k^z}) = \sum_{i=0}^{n_2-1} \mathbf{k}^{\mathbf{z}}_{m,\ell,i} \cdot \text{IDFT}_i(t)\\
\end{equation}
where $\mathbf{k}^{\mathbf{z}}_{m,\ell,i} \in \mathbb{R}^3 $ is defined as $\mathbf{k}^{\mathbf{z}} = (k_{m,\ell,i})^{m:-\ell\leq m \leq \ell}_{\ell: 0\leq \ell \leq \ell_{max}}$ and $0 \leq i \leq n_2$ is an additional dimension to store corresponding Fourier coefficients for each SH element $z_{m,l}$.

Similar to PlenOctree~\cite{yu2021plenoctrees}, the mapping function $\Phi$ can be adapted by an MLP based NeRF network which we called Fourier NeRF-SH and be further discretized into octree-based volume representation.
Content in each leaf contains the Fourier coefficients $\mathbf{k}^\sigma$ and  $\mathbf{k^z}$ of the corresponding position. 
As a result, the proposed representation absorbs the advantages and benefits of the original PlenOctree and enables real-time novel view synthesis for free-viewpoint videos.

\begin{figure}[t]
\begin{center}
    \includegraphics[width=\linewidth]{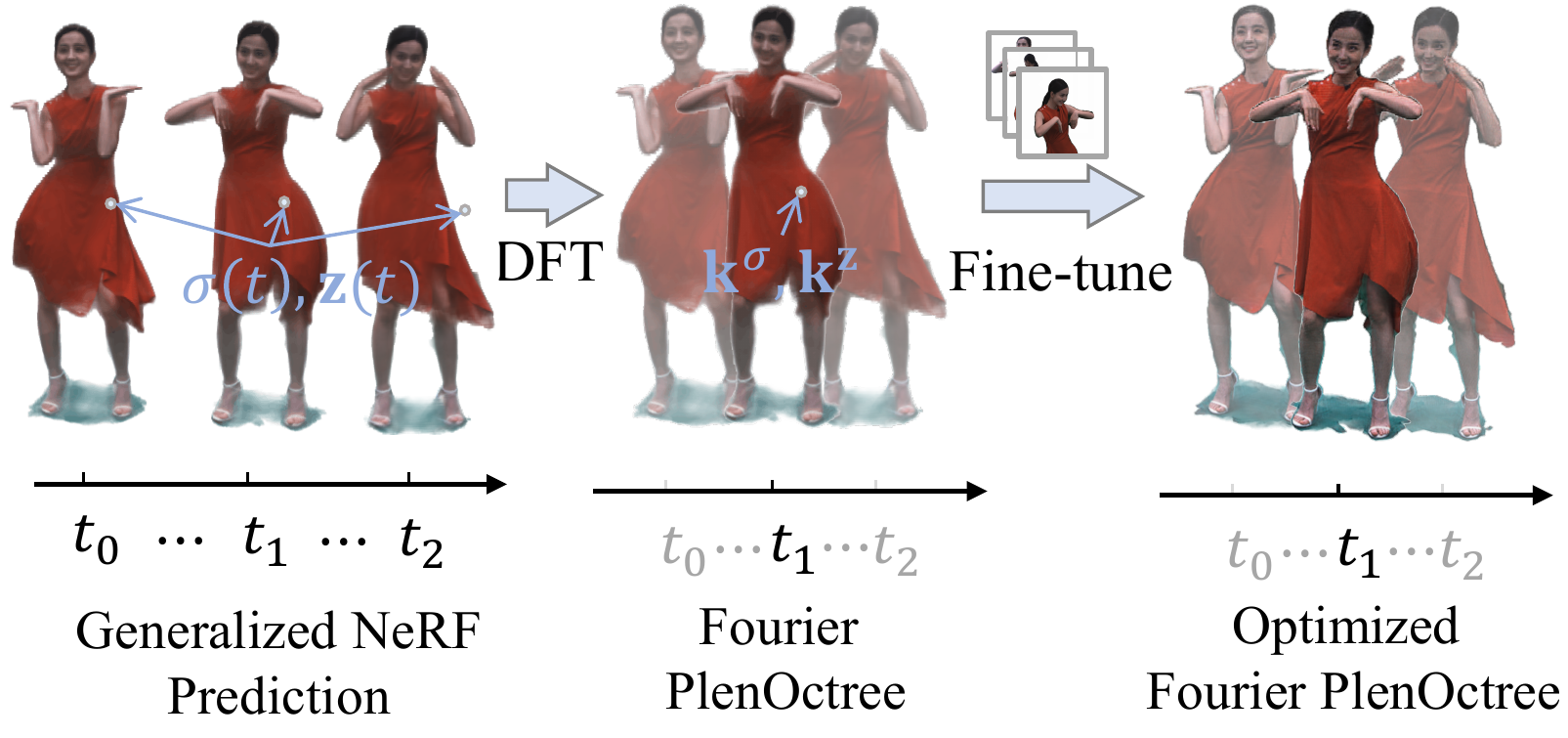}
\end{center}
\vspace{-6mm}
\caption{A straightforward pipeline to generate Fourier PlenOctree from multiple PlenOctrees, for each leaf in Fourier PlenOctree, we first find all corresponding PlenOctree leaves at the same position. They store a sequence of density $\sigma$ and SH coefficients $\mathbf{z}(t)$ along the time axis. We can convert them to Fourier coefficients $\mathbf{k}^\sigma$ and $\mathbf{k^z}$ corresponding to density and SH coefficients, and store them in Fourier PlenOctree. Finally, we can optimize Fourier PlenOctree using ground truth images.}
\label{fig:fig4}
\vspace{-6mm}
\end{figure}

\subsection{Fourier PlenOctree Generation}\label{sec:4.2}

Reconstruction a Fourier PlenOctree as described in Sec.~\ref{sec:4.1} is a big challenge. 
A naive way to reconstruct such Fourier PlenOctree is to fit a continual implicit function as described in Eq.~\ref{eqn:representation} from scratch using Fourier NeRF-SH like~\cite{yu2021plenoctrees}, which takes about 1-2 days. 
For speed considerations, we adopt Generalized PlenOctree Fusion (Section.~\ref{sec:3}) in free-viewpoint video generation.

Octree structures vary from frame to frame due to object's motion. 
Fourier PlenOctree requires the structures to be the same in all frames in order to analyze plenoptic functions located at the same position.  
Fortunately, we are able to fast infer octree structures via Shape-From-Silhouette (SFS).
Applying Generalized PlenOctree Fusion for each frame, we fill content in all frames' PlenOctrees.
After that, all we need is to unify them.
For PlenOctrees at timestamps $t = 1, 2, \cdots, T$, we first calculate the union of their structures, note that the union of their structures always has equal or deeper leaves comparing PlenOctree at any frame. 
In other words, each leaf in an octree either is divided or keeps the same. 
In the case of division, we just copy the pre-existed value from the parent node (previous leaf) to new leaves. 

\begin{figure*}[t]
\begin{center}
    \includegraphics[width=0.95\linewidth]{ 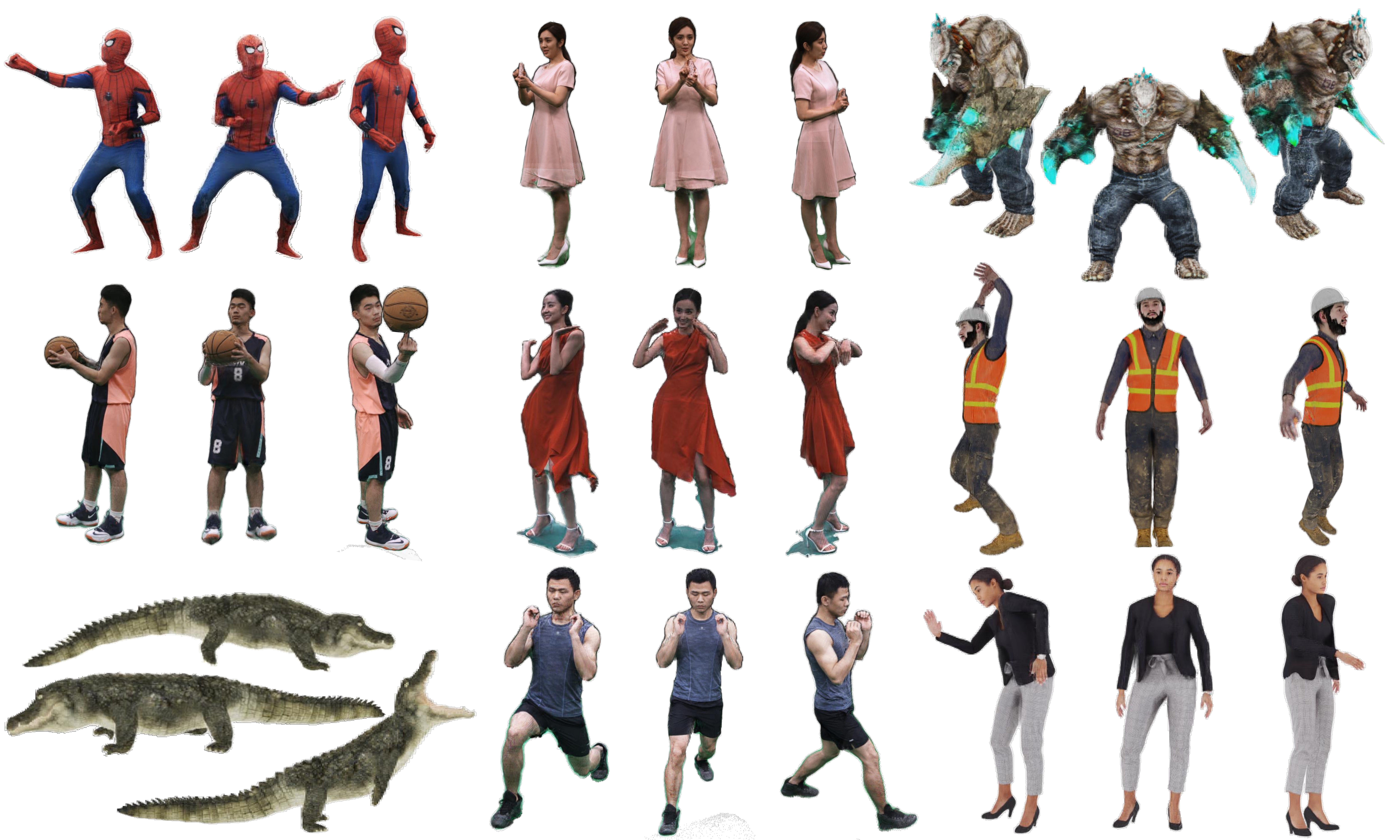}
\end{center}
\vspace{-0.5cm}
\caption{The rendered appearance results of our Fourier PlenOctree method on several sequences.}
\label{fig:gallery}
\vspace{-5mm}
\end{figure*}

Having unified Plenoctrees for each frame, we calculate a Fourier PlenOctree, as shown in Fig. \ref{fig:fig4}, which has the same octree structure as theirs by using the Discrete Fourier Transform (DFT) for each leaf's values $\mathbf{k}^\sigma $ and $ \mathbf{k^z} $:
\begin{equation} \label{eqn:dft_sigma}
    \mathbf{k}^\sigma_i = \sum^{T}_{t=1} \sigma(t) \cdot \text{DFT}_i(t)
\end{equation}
\begin{equation} \label{eqn:dft_z}
    \mathbf{k}^{\mathbf{z}}_{m,\ell,i} =  
    \sum^{T}_{t=1} z_{m,\ell}(t) \cdot \text{DFT}_i(t)
\end{equation}
where
\begin{equation}
    \text{DFT}_i(t) = 
    \left\{
    \begin{aligned}
    &\frac{1}{T}\cos(\frac{i\pi}{T} t) &\text{ if $i$ is even}\\
    &\frac{1}{T}\sin(\frac{(i+1)\pi}{T} t) &\text{ if $i$ is odd}\\
    \end{aligned}
    \right.
\end{equation}

\subsection{Fourier PlenOctree Fine-tuning} \label{sec:4.3}
Although our proposed Fourier PlenOctree has a DFT mapping function from Fourier coefficients to densities and SH coefficients at a specific timestamp, the fine-tuning procedure discussed in~\cite{yu2021plenoctrees} can be extended to our method to improve the image quality via back propagation as DFT is totally differentiable.

The objective function of this procedure is the same as the loss function of \cite{mildenhall2020nerf}:

\begin{equation}
    \mathcal{L} = \sum_{t=1}^T\sum_{i=1}^N\|\hat{I}^t_i-I^t_i\|_2^2
\end{equation}
where $\hat{I}^t_i$ is the rendered image for view $i$ and frame $t$.

The optimization time is much shorter than optimizing Fourier NeRF-SH, since Fourier PlenOctree is an explicit representation which is easier to optimize than MLP-based implicit representations.

\section{Experimental Results}

In this section, we evaluate our Fourier Plenoctree method on a variety of challenging scenarios. We run our experiments on a PC  with a single NVIDIA GeForce RTX3090 GPU. It only takes us about 2 hours to reconstruct Fourier PlenOctree with input from ~60 views and ~60 frames.  For dynamic datasets, we have five real datasets in $2048\times1536$ and five synthetic datasets in $1080\times1080$. We use $\ell_{max}=2$ (9 components) and $512^3$ grid size for our Fourier PlenOctree. Our method achieves rendering speed at 100fps with 800$\times$800 resolution, which is ~3000 times faster than the original NeRF. As demonstrated in Fig.~\ref{fig:gallery}, our approach generates high-quality appearance results and even handles identities with rich textures and challenging motions. Please refer to the supplementary video for more video results.

\begin{figure*}[t]
\begin{center}
    \includegraphics[width=\linewidth]{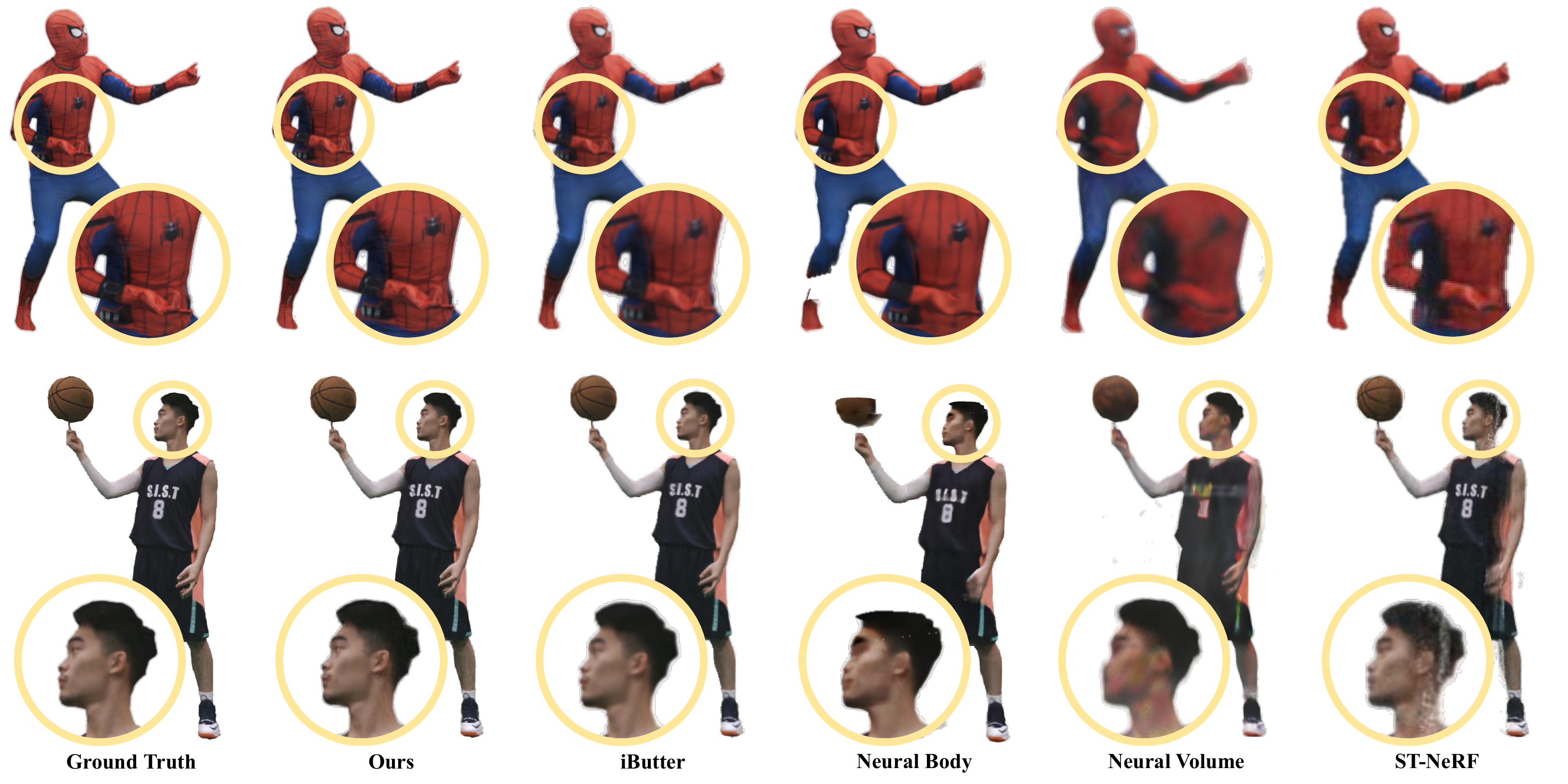}
\end{center}
\vspace{-0.5cm}
\caption{Qualitative comparison against dynamic scene rendering methods. We compare our method with Neural Body, Neural volumes and ST-NeRF and our approach generalizes the most photo-realistic and finer detail.}
\label{fig:comparison}
\vspace{-3mm}
\end{figure*}

\subsection{Comparison}

Our approach is first to enable the fast generation and real-time dynamic rendering to the best of our knowledge.
To demonstrate the overall performance of our approach, we compare to the existing free-viewpoint video methods based on neural rendering, including the voxel-based method \textbf{Neural Volumes}~\cite{lombardi2019neuralvol}, and implicit methods \textbf{iButter}~\cite{wang2021ibutter}, \textbf{ST-NeRF}~\cite{st-nerf} and \textbf{Neural Body}~\cite{peng2021neural} based on neural radiance field. For a fair comparison, all the methods share the same training dataset as our approach.

As shown in Fig.~\ref{fig:comparison}, our approach achieves photo-realistic free-viewpoint rendering with the most vivid rendering result in terms of photo-realism and sharpness, which, in addition, can be done in real-time. 

For quantitative comparison, we adopt the peak signal-to-noise ratio (\textbf{PSNR}), structural similarity index (\textbf{SSIM}), mean absolute error (\textbf{MAE}), and Learned Perceptual Image Patch Similarity (\textbf{LPIPS}) \cite{zhang2018perceptual} as metrics to evaluate our rendering accuracy. 
We keep 90 percent of captured views as training set and the other 10 percent views as testing set.
As shown in Tab.~\ref{table:quantitative comparison_perform}, our approach outperforms other methods in terms of all the metrics for appearance.
Such a qualitative comparison illustrates the effectiveness of our approach to encode the spatial and temporal information from our multi-view setting.
In Tab.~\ref{table:quantitative comparison_time}, our method achieves the fastest rendering in dynamic scenes and uses the second least training or fine-tuning time given a new multi-view sequence.

\begin{table}[t]
	\centering
	\small{
\begin{tabular}{l|c|c|c|c}
	\multicolumn{5}{c}{ \colorbox{best1}{best} \colorbox{best2}{second-best} } \\
	Method        &  PSNR$\uparrow$        & SSIM$\uparrow$          &MAE$\downarrow$          &LPIPS$\downarrow$  \\ \hline
	Neural Body   & 27.34                  & 0.9414              &  0.0123                & 0.0373\\
	NeuralVolumes & 23.62                  & 0.9219                  & 0.0251                  & 0.0881   \\
	ST-NeRF       & 30.63                 & 0.9486               & 0.0092              & 0.0324\\
	iButter       & \cellcolor{best2}33.77 & \cellcolor{best2}0.9596 & \cellcolor{best2}0.0054 & \cellcolor{best2}0.0295\\ \hline
	Ours          & \cellcolor{best1}35.21 & \cellcolor{best1}0.9910 & \cellcolor{best1}0.0033 & \cellcolor{best1}0.0217 \\    \hline
\end{tabular}
    }
\rule{0pt}{0.05pt}
\caption{\textbf{Quantitative comparison against several methods in terms of rendering accuracy.} Compared with ST-NeRF, Neural Volumes, NeuralBody and iButter , our approach achieves the best performance in \textbf{PSNR}, \textbf{SSIM}, \textbf{LPIPS} and \textbf{MAE} metrics.}
\label{table:quantitative comparison_perform}
\vspace{-1mm}
\end{table}

\begin{table}[t]
	\centering
	\small{
	\begin{tabular}{l|c|c}
	\multicolumn{3}{c}{ \colorbox{best1}{best} \colorbox{best2}{second-best} } \\
        Method        & Time$\downarrow$ & FPS$\uparrow$ \\ \hline
        Neural Body   &  9.6 hours                & 0.34\\
		NeuralVolumes &  6 hours                  & 2.3 \\
		ST-NeRF       & 12 hours                  & 0.04 \\
		iButter       & \cellcolor{best1}20 mins  & \cellcolor{best2}3.54\\ \hline
		Ours          & \cellcolor{best2}2 hours  & \cellcolor{best1}100 \\    \hline
    \end{tabular}
    }
\rule{0pt}{0.05pt}
\caption{\textbf{Quantitative comparison against several methods in terms of  training and rendering speed.} Compared with NeuralBody, Neural Volumes, ST-NeRF and iButter, our approach achieves the best performance in FPS metrics and the second best in training or fine-tuning time.}
\label{table:quantitative comparison_time}
\vspace{-5mm}
\end{table}

\subsection{Ablation Study}

\begin{figure}[t]
\begin{center}
    \includegraphics[width=0.98\linewidth]{ 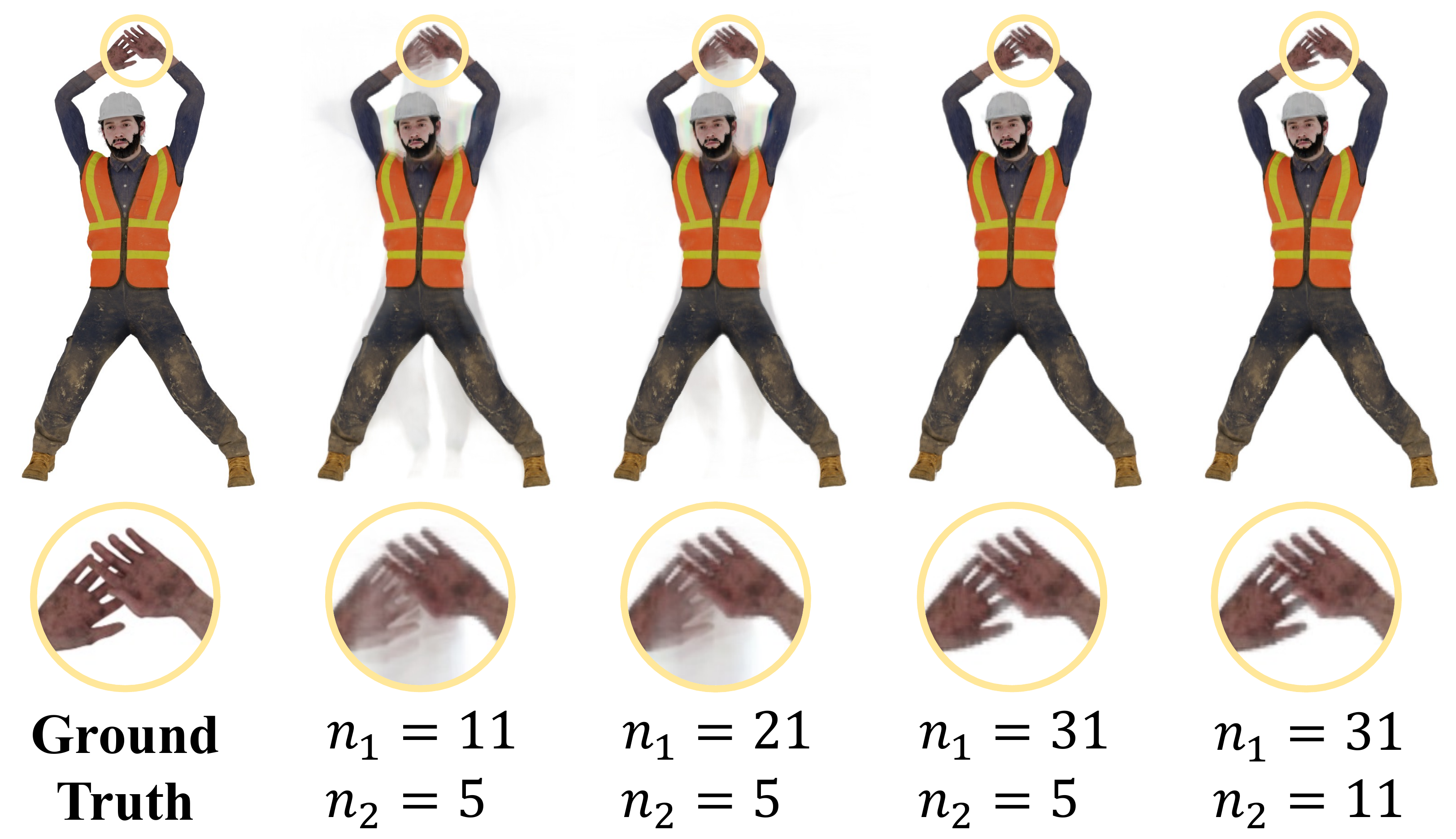}
\end{center}
\vspace{-0.5cm}
\caption{\textbf{Qualitative evaluation on Fourier dimensions.} The setting with $n_1=31$, $n_2=5$ achieves the satisfactory rendering quality while higher Fourier dimension does not result in a significant improvement.}
\label{fig:eval1}
\vspace{-5mm}
\end{figure}

\textbf{Fourier dimensions.} We carried out an experiment to find the best choice of Fourier dimension with both realistic rendering performance and acceptable memory usage. As is shown in Fig. \ref{fig:eval1} and Tab. \ref{table:eval1}, the results with $n_1=31$, $n_2=5$ have a better appearance than those using smaller Fourier dimensions and have less storage cost and faster rendering than using higher dimensions. Our model keeps an outstanding balance. 

\begin{figure}[t]
\begin{center}
    \includegraphics[width=0.98\linewidth]{ 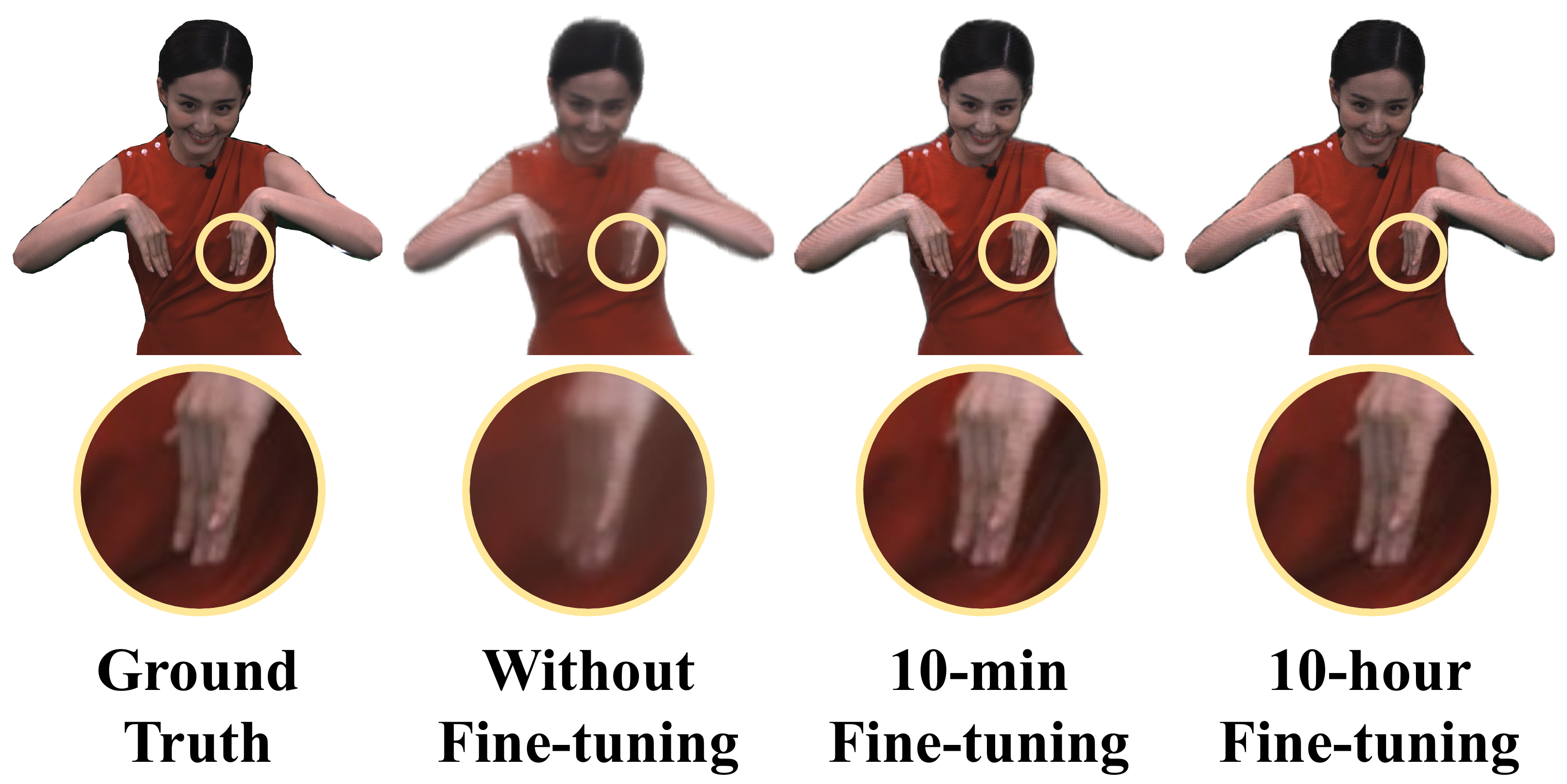}
\end{center}
\vspace{-0.3cm}
\caption{\textbf{Qualitative evaluation on time of fine-tuning.} 10-minute fine-tuning achieves the Satisfactory rendering quality while longer fine-tuning does not result in a significant improvement.}
\label{fig:eval2}
\vspace{-3mm}
\end{figure}

\textbf{Time of fine-tuning.} We analyze the relationship between rendering quality and fine-tuning time. In these experiments, we compared the results obtained without fine-tuning, with a 10-minute fine-tuning, and with a 10-hour fine-tuning which is long enough. As is shown in Fig. \ref{fig:eval2}, the results without fine-tuning have blured in the details. After 10 minutes of fine-tuning, such artifacts were removed. We also found that further fine-tuning only turns out slight improvement. Quantitative results in Tab. \ref{table:eval2} shows the same phenomenon.

\textbf{4D Representation in Frequency Domain.} We also evaluate the efficiencies of our 4D Representation variants from the perspectives of rendering quality, storage, and consumed time. 
We set a limit of memory usage as 24GB in our experiments. %

\begin{table}[t]
	\centering
	\resizebox{\linewidth}{!}{ %
    	\begin{tabular}{l|c|c|c}
	    \multicolumn{4}{c}{ \colorbox{best1}{best} \colorbox{best2}{second-best} } \\
            Fourier dimensions     & PSNR$\uparrow$         & FPS $\uparrow$          & Storage (GB)$\downarrow$ \\ \hline
            $n_1=11~ n_2=5$        & 31.56                  & \cellcolor{best1}118.47 & \cellcolor{best1}6.421\\
    		$n_1=21~ n_2=5$        & 33.31                  & \cellcolor{best2}118.14 & \cellcolor{best2}6.861  \\
    		$n_1=31~ n_2=5$ (ours) & \cellcolor{best2}36.21 & 117.87                  & 7.251  \\
    		$n_1=31~ n_2=11$       & \cellcolor{best1}36.40 & 109.95                  & 14.91\\  \hline
        \end{tabular}
    }
\rule{0pt}{0.05pt}
\caption{\textbf{Quantitative evaluation on Fourier dimensions.} Compared with other choices, the setting with $n_1=31$, $n_2=5$ achieves the best balance among rendering accuracy, time and storage.}
\label{table:eval1}
\end{table}

\begin{table}[t]
    
	\centering
	\resizebox{\linewidth}{!}{ %
    	\begin{tabular}{l|c|c|c|c}
    	    \multicolumn{5}{c}{ \colorbox{best1}{best} \colorbox{best2}{second-best} } \\
            Method                    & PSNR$\uparrow$          & SSIM$\uparrow$          & MAE$\downarrow$         & LPIPS$\downarrow$ \\ \hline
            w/o fine-tuning           & 26.02                   & 0.9671                  & 0.0126                  & 0.0678 \\
            10-min fine-tuning (ours) & \cellcolor{best2}32.93  & \cellcolor{best1}0.9766 & \cellcolor{best2}0.0050 & \cellcolor{best2}0.0340 \\
        	10-hour fine-tuning       & \cellcolor{best1}33.39  & \cellcolor{best2}0.9763 & \cellcolor{best1}0.0046 & \cellcolor{best1}0.0336 \\
        \end{tabular}
        }
\rule{0pt}{0.05pt}
\caption{\textbf{Qualitative evaluation on time of fine-tuning.} 10-minute fine-tuning achieves the Satisfactory rendering quality while 10-hour fine-tuning does not result in a significant improvement.}
\label{table:eval2}
\end{table}

\begin{table}[t]
	\centering
	\resizebox{\linewidth}{!}{ %
    	\begin{tabular}{c|c|c|c|c}
	    \multicolumn{5}{c}{ \colorbox{best1}{best} \colorbox{best2}{second-best} } \\
            Method &  PSNR$\uparrow$        & GB$\downarrow$         & Realtime Rendering & Fine-tuning time$\downarrow$ \\ \hline
            \textbf{(a)}    & \cellcolor{best2}32.15 & \cellcolor{best1}7.033 & \Checkmark         & \cellcolor{best1}2 hours \\ 
            \textbf{(b)}    & 25.97                  & \cellcolor{best2}8.669 & \Checkmark         & \cellcolor{best2}10 hours \\
	        \textbf{(c)}    & \cellcolor{best1}32.39 & 74.32                  & \XSolidBrush       & 19 hours \\
        \end{tabular}
    }
\rule{0pt}{0.05pt}
\caption{\textbf{Quantitative evaluation on our Fourier Plenoctree.} \textbf{(a)} Our model, using Fourier representation, limited memory   
w/ DFT (ours),\textbf{ (b)} w/o DFT, limited memory,  \textbf{(c)} w/o DFT, unlimited memory. Our model is able to use minimal storage with the least fine-tuning time to achieve high fidelity results. 
}
\label{table:eval3}
\vspace{-5mm}
\end{table}

\begin{figure}[t]
\begin{center}
    \includegraphics[width=0.95\linewidth]{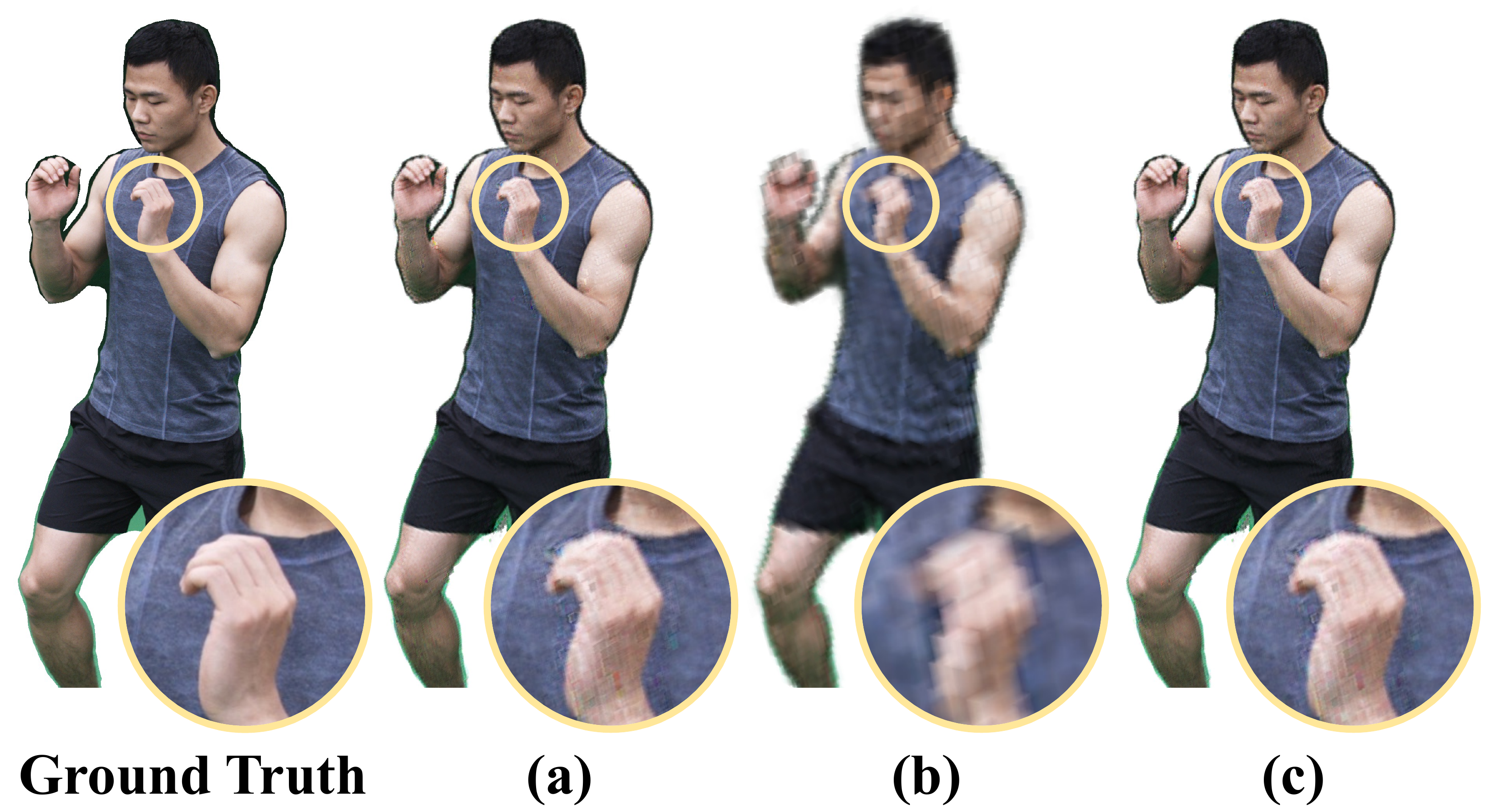}
\end{center}
\vspace{-0.3cm}
\caption{\textbf{Qualitative evaluation on DFT.}\textbf{ (a)} Our model, using Fourier representation, limited memory   
w/ DFT, limited memory (ours), \textbf{(b)} w/o DFT, limited memory,  \textbf{(c)} w/o DFT, unlimited memory. Our model is able to use minimal storage with the least training time to achieve high fidelity results. The use of DFT can significantly improve the quality of the results.}
\label{fig:eval3}
\vspace{-3mm}
\end{figure}

As is shown in Fig. \ref{fig:eval3} and Tab. \ref{table:eval3}, when the memory is limited, the use of DFT can significantly improve the quality of the results to the case where there is no memory limitation. Also, our model uses the least storage and training time to enable dynamic real-time rendering compared to other methods.

\section{Discussion}

\textbf{Limitation.}
As the first trial to enable fast generation of octree-based representations and real-time rendering for both static and dynamic scenes, our approach has some limitations.

First, despite using a generalized NeRF to directly predict density and color for scenes from input images, we still need dense inputs for static or dynamic scenes.  The capturing settings are still expensive and hard to construct. Second, compared with implicit representations such as MLP-based representation for static or dynamic scenes, Fourier PlenOctree still needs larger storage and GPU memory requirement. Also, when the motion of the dynamic scene is faster or the frame number of the multi-view videos is more extended, a higher dimension of Fourier coefficients is needed to keep a high quality of rendered results, which requires higher storage and GPU memory. Finally, we cannot handle large movements of entities like walking performers on the street. Our approach is inefficient as we use the union of visual hulls to initialize Fourier PlenOctrees. 

\textbf{Conclusion.}
We have presented a novel Fourier PlenOctree (FPO) technique for efficient neural modeling and real-time rendering of dynamic scenes captured under the free-view video setting.
Our coarse-to-fine fusion scheme combines generalizable NeRF with PlenOctree for efficient neural scene construction in minutes.
We construct the FPO by tailoring the implicit network to model Fourier coefficients, achieving high-quality rendering for dynamic objects in real-time with compact memory overload. 
Our experimental results demonstrate the effectiveness and efficiency of FPO for high-quality dynamic scene modeling.
With the unique fast generation and real-time rendering ability for dynamic scenes, we believe that our approach serve as a critical step for neural scene modeling, with various potential applications in VR/AR and immersive telepresence. 
{\small
\bibliographystyle{ieee_fullname}
\bibliography{egbib}
}

\end{document}